\documentclass[10pt,twocolumn,letterpaper]{article}

\usepackage{wacv}
\usepackage{times}
\usepackage{epsfig}
\usepackage{graphicx}
\usepackage{amsmath}
\usepackage{amssymb}
\usepackage{booktabs}
\usepackage{textcomp}

\def\wacvPaperID{145} 

\wacvfinalcopy 

\ifwacvfinal
\def\assignedStartPage{1} 
\fi

\ifwacvfinal
\usepackage[breaklinks=true,bookmarks=false]{hyperref}
\else
\usepackage[pagebackref=true,breaklinks=true,colorlinks,bookmarks=false]{hyperref}
\fi

\ifwacvfinal
\else
\fi

\begin{document}

\title{AWADA: Attention-Weighted Adversarial Domain Adaptation for Object Detection}

\author{Maximilian Menke\\
Robert Bosch GmbH\\
{\tt\small maximilian.menke@de.bosch.com}
\and
Thomas Wenzel\\
Robert Bosch GmbH\\
{\tt\small thomas.wenzel2@de.bosch.com}
\and
Andreas Schwung\\
Fachhochschule Südwestfalen\\
{\tt\small schwung.andreas@fh-sef.de}
}

\maketitle

\begin{abstract}
   Object detection networks have reached an impressive performance level, yet a lack of suitable data in specific applications often limits it in practice. Typically, additional data sources are utilized to support the training task. In these, however, domain gaps between different data sources pose a challenge in deep learning. GAN-based image-to-image style-transfer is commonly applied to shrink the domain gap, but is unstable and decoupled from the object detection task. \\
   We propose AWADA, an Attention-Weighted Adversarial Domain Adaptation framework for creating a feedback loop between style-transformation and detection task. By constructing foreground object attention maps from object detector proposals, we focus the transformation on foreground object regions and stabilize style-transfer training. \\
   In extensive experiments and ablation studies, we show that AWADA reaches state-of-the-art unsupervised domain adaptation object detection performance in the commonly used benchmarks for tasks such as synthetic-to-real, adverse weather and cross-camera adaptation.
\end{abstract}

\section{Introduction}

In deep learning, extensive real-world data collection aims at creating large training datasets for downstream tasks such as object detection. Labeling such data can be very expensive because, typically, human annotators have to annotate each object instance by hand. \\
One direction of research focuses on unsupervised domain adaptation methods, which transform knowledge learned from a labeled source domain to a related unlabeled target domain. Hence, no labels on the target domain are required anymore. Object detectors typically suffer on a target domain when trained on a source domain because of the gap between both data domains. In domain adaptation, by utilizing Generative Adversarial Networks (GANs) \cite{goodfellow2014generative}, current style-transfer methods transform the image style of a source domain to a target domain style. Therefore, the style-transfer network aligns the data distributions on the image level. \\
Especially for object detection, foreground object regions are more relevant than the background. Current style-transfer networks based on e.g. Cycle-GAN \cite{CycleGAN} ignore this foreground-background distinction and use all pixels equally weighted to solve generators and discriminators' adversarial min-max optimization problems. Therefore, the style-transfer network is disconnected from the final downstream task, i.e. object detection. \\
We propose AWADA, an Attention-Weighted Adversarial Domain Adaptation framework for object detection, which targets this problem. Using foreground and background distinction based on object proposals (see Figure \ref{fig:AttentionMaps}) we re-weight the style-transfer loss functions using our proposed Attention-Weighting Modules (AWM). Our method restricts the adversarial game between generators and discriminators focusing on object foreground regions. With AWADA, we create a novel feedback loop between downstream task and style-transfer network to target disconnected transformation and detection tasks we see in current SOTA style-transfer networks. Our method also stabilizes the training of the GAN-based style-transfer training. \\
We evaluate AWADA on different automotive benchmarks as synthetic-to-real, adverse weather, and cross-camera adaptation. We show that AWADA outperforms current unsupervised domain adaptation methods. Our contributions are the following:

\begin{figure*}[t]
	\centering
	\includegraphics[width=0.9\textwidth]{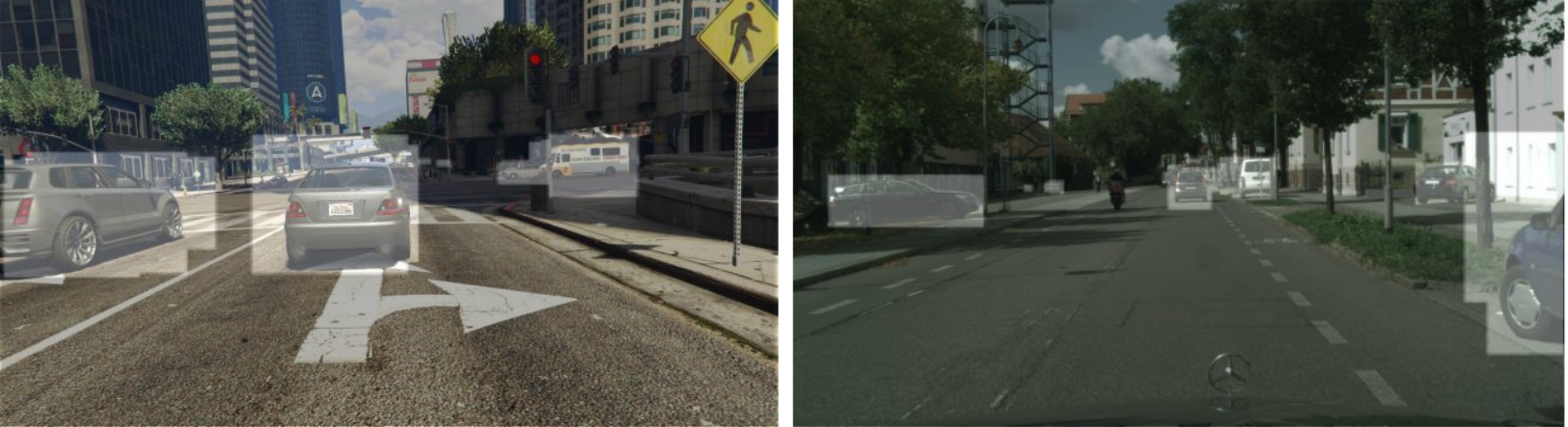}
	\caption{Exemplary images and corresponding attention maps constructed from object detector proposals of sim10k \cite{sim10k} on the left and Cityscapes \cite{Cityscapes} on the right. Bright overlays mark attention regions of foreground object used to re-weight our GAN losses. Note that only car objects are used to construct attention maps for both images.}
	\label{fig:AttentionMaps}
\end{figure*}

\begin{itemize}
	\item We propose AWADA, a novel style-transformation framework, which introduces a feedback loop between the object detection and the transformation task. 
	\item We weight foreground and background areas differently when training our style-transfer network. AWADA focuses the adversarial training on object regions using attention maps constructed from object detector proposals.
	\item We show that AWADA outperforms current GAN-based SOTA methods in unsupervised domain adaptation on three standard automotive benchmarks.
\end{itemize}

\section{Related Work}

This section gives an overview of current state-of-the-art methods in unsupervised domain adaptation. We first give a broad overview of domain adaptation in object detection. Afterwards we discuss GAN-based style-transfer networks, which are primarily motivated by semantic segmentation. Finally, we outline current trends in utilizing foreground-background distinction in domain adaptation. 

\subsection{Domain Adaptation in Object Detection}
Domain adaptation in deep learning object detection started in 2018 by introducing global and local gradient reversal layers (GRL \cite{GRL}) into a Faster-RCNN object detector network \cite{DAFasterRCNN,StrongWeak}. Recent work utilizes image and instance full alignment (iFAN) to align multi-level features between two related data domains \cite{iFAN}. The authors in \cite{SeekingSimilarities} propose a similarity-based domain alignment by matching only features of objects belonging to the same cluster. Furthermore, normalizing features of two domains before applying GRL is discussed by Su et al. \cite{CDN} and suppression of gradients is applied by Wang et al. \cite{wang2021domain}. In the direction of feature learning, there are several other publications \cite{xu2020exploring,chen2020harmonizing,vs2021mega}.\\
Another research direction e.g. by the authors of \cite{ProgressiveDomainAdaptation}, \cite{li2021category} and \cite{UDA} leverage iterative pseudo labeling of the target domain. Work by Soviany et al. uses curriculum learning \cite{Curriculum} by first creating pseudo labels for easy instances and iteratively takes more difficult instances into account. In \cite{UMT}, the authors applied Unbiased Mean Teacher, which uses knowledge distillation to align two domains in a student-teacher setup. \\
With DA-DETR \cite{DADETR}, transformers were applied in domain adaptation for the first time. Using hybrid self-attention modules, they improve object detection performance while also providing a strong transformer baseline across different automotive benchmarks. Another research direction is using Graph Neural Networks (GNN) to close the domain gap by structuring the domain distributions using graphs \cite{xu2020cross,li2022scan}. \\
Contrary to other approaches, we do not modify the Faster-RCNN object detector from Ren et al. \cite{FasterRCNN}. Therefore our work is orthogonal to other's work \cite{DAFasterRCNN,EveryPixelMatters,iFAN}, as we keep the detector network untouched and apply style-transfer to align the data distributions.

\subsection{GAN-based Domain Adaptation}

\begin{figure*}[t]
	\centering
	\includegraphics[width=0.9\textwidth]{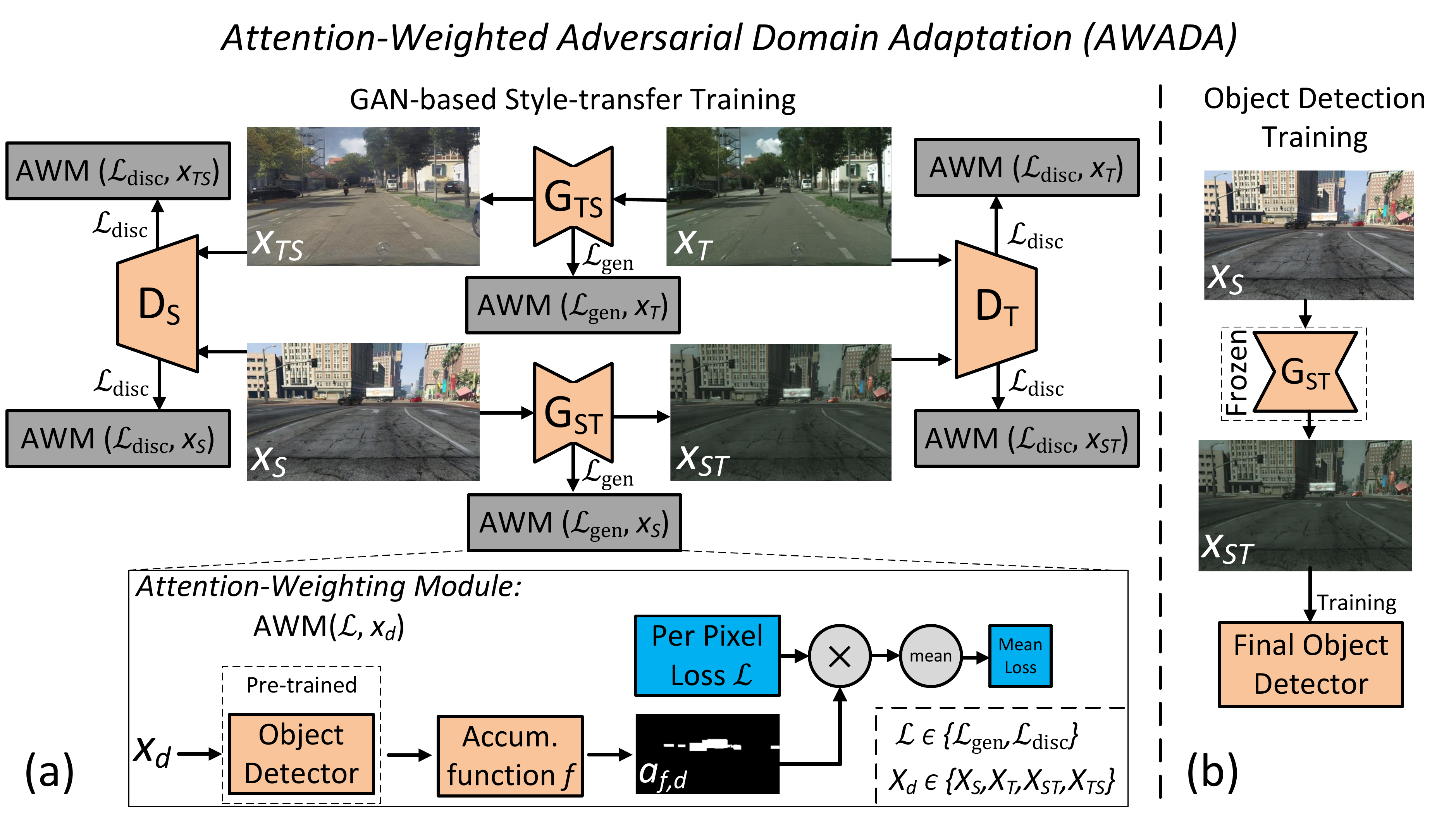}
	\caption{In (a), we show our proposed AWADA framework with training of source to target generator $G_{ST}$, target to source generator $G_{TS}$ and two discriminators $D_{S}$ and $D_{T}$. With our proposed Attention-Weighting Modules (AWM) we utilize foreground object attention maps, drawn from pre-trained object detectors. Source domain attention maps are obtained from a source domain-trained detector, while the target domain detector is trained on the stylized source domain. We re-weight the per-pixel loss of each adversarial component of the style-transfer network using our attention maps. Afterwards in (b), we train a final object detector on AWADA stylized images while freezing the generator $G_{ST}$.}
	\label{fig:Framework}
\end{figure*}

In semantic segmentation, the first approach suing style-transfer for domain adaptation was Cycle-GAN \cite{CycleGAN}, introducing cycle-consistency to enforce consistent generation with reducing visual artifacts. CyCADA \cite{CyCADA} introduced semantic consistency into the Cycle-GAN framework. MADAN \cite{MADAN} extended CyCADA by utilizing multiple source domains in their generation framework. Other orthogonal style-transfer frameworks as UNIT \cite{UNIT} and Pix2PixHD \cite{pix2pixHD} have been released as well. \\
Style-transfer has also been applied to object detection by a wide range of other publications \cite{ProgressiveDomainAdaptation,UDA,Curriculum}. They mostly rely on a fixed Cycle-GAN style-transfer network to transform the image style of the source domain. However, the authors of \cite{menke2022} propose to reduce the patch size in training and adapt CyCADA from semantic segmentation to object detection. They have shown, that including semantic consistency in style transfer also improves object detection performance. \\
We extend this approach by introducing object detection-guided attention into the transformation cycle to modify the style-transfer network specifically for object detection, which has not been considered in related work before.

\subsection{Attention in Domain Adaptation}
Leveraging foreground object attention mechanisms in domain adaptation is not new anymore. Zheng et al. propose a method guiding the GRL-based domain transfer by attention maps drawn from an object detector \cite{CTF}. Specific for anchor-less object detection models \cite{EveryPixelMatters} utilizes center and objectiveness scores to align intermediate feature representations. Yang et al. weight multi-level GRL predictions with hard-attention maps constructed from object proposals \cite{FFDA}. By utilizing pseudo attention maps acquired from box-level semantic segmentation, the authors of \cite{C2FDA} propose a course domain alignment framework. Using self-attention on multi-level feature maps, ILLUME \cite{ILLUME} guides cross-domain feature alignment with self-attentive gradient reversal layers. \\
Previous methods try to perform selective feature alignment based on attention mechanisms. In contrast, we combine attention with style-transfer by creating a feedback loop from object detection to the style-transfer network. We note that previous research using style-transfer does not connect the downstream task to the style-transfer network. Therefore we propose to use attention maps drawn from object proposals to re-weight the loss functions of style-transfer networks.

\section{Proposed Method}
In this section, we describe our proposed method AWADA. First, we define the standard unsupervised domain adaptation setting for object detection. Then, we explain our AWADA framework and show how we combine it with object detection. Later, we describe how we construct attention maps from an object detector and use them to re-weight loss functions of style-transfer networks using our proposed Attention-Weighting Modules (AWM) to create our proposed AWADA framework.

\subsection{Unsupervised Domain Adaptation in Object Detection}
We assume a labeled source domain $\mathbb{D}_S$ containing source images $X_S = \{x_S^i\}_{i = 1}^{N_S}$ and corresponding bounding box labels $Y_S = \{y_S^i\}_{i = 1}^{N_S}$. We further assume an unlabeled target domain $\mathbb{D}_T$ with target images $X_T = \{x_T^i\}_{i = 1}^{N_T}$ and no labels $Y_T = \{\}$. The number of samples in each domain is represented by $N_S$ for the source domain and $N_T$ for the target domain respectively. Our setting is unsupervised in the target domain, as we do not use any labels from it except for evaluation purposes. \\
With unsupervised domain adaptation, our aim is to extract predictions $\hat{Y}_T = \{{\hat{y}_{T}}^i\}_{i = 1}^{N_T}$ from an object detector trained without target domain labels. The two domains should be related to each other regarding class occurrences.

\begin{figure*}[t]
	\centering
	\includegraphics[width=0.9\textwidth]{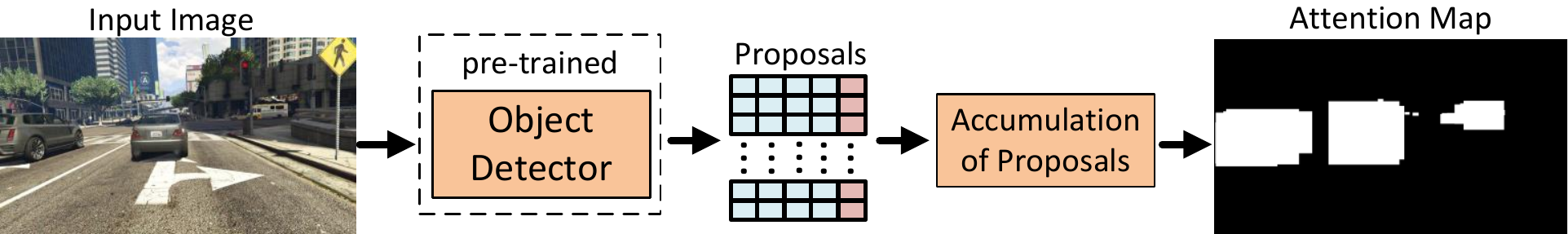}
	\caption{An input image is fed through a pre-trained object detector to create foreground object proposals. For source images we use a detector trained on the source domain and for target images we use a detector trained on the stylized source domain. Each proposal has a four-dimensional box descriptor and a confidence score. We create attention maps of input image size by accumulating proposals per spatial position.}
	\label{fig:Sampling}
\end{figure*}

\subsection{Framework Overview}

Our proposed AWADA framework in Figure \ref{fig:Framework} is based on CyCADA* \cite{menke2022}. We extend on CyCADA* by adding our Attention-Weighting Modules (AWMs) to each generator and discriminator loss to focus the style-transfer network training on foreground objects. AWADA consists of a generator transforming source images to target-style images. Additionally, there is a generator transforming target images to source-like images. Our framework also consists of two discriminators that predict the root domain of each image. \\
Furthermore we define a novel training sequence for domain adaptive object detection. First, we train the CyCADA* style-transfer network without foreground-background distinction. Afterwards we train a Faster-RCNN \cite{FasterRCNN} object detector on CyCADA* stylized source images $X_{ST}$ to construct attention maps for the target domain. We also train an object detector on the source images to construct attention maps for the source domain. Finally, we re-weight adversarial loss functions with the attention maps using our proposed AWMs and train AWADA to focus the transformation on foreground object regions. Afterwards we freeze the source-to-target generator of AWADA to transfer source images to the target style. Using these stylized source images, we train our final Faster-RCNN object detector and evaluate on the target domain. \\
By utilizing attention maps constructed from pre-trained object detector proposals, we are the first to connect the downstream task with the style-transfer network. Therefore, our proposed AWADA framework constructs a feedback loop between the object detection downstream task and the style-transfer network. Following we explain each individual components of our AWADA framework in more detail.

\subsection{Style-Transfer Baseline CyCADA*}
We follow the authors of \cite{menke2022} and adapt CyCADA \cite{CyCADA} from semantic segmentation to object detection, and name their approach CyCADA*. In this section we describe some details of the latter, since our proposed AWADA framework is based on it.\\
Style-transfer networks based on Cycle-GAN \cite{CycleGAN} rely on two generators and two discriminators. The generators should generate high-quality fake samples, which the discriminators classify as such. A generator $G_{ST}$ transforms source image patches into target-like ones. Contrary, a second generator $G_{TS}$ transforms target images into source-like ones. The discriminators $D_S$ and $D_T$ use adversarial GAN loss $\mathcal{L}_{GAN}^{ST}$ and $\mathcal{L}_{GAN}^{TS}$ respectively to predict the root domain:

\begin{multline}
	\label{eq:GAN1}
	\mathcal{L}_{GAN}^{ST}(G_{ST}, D_T, x_S, x_T) = \\
	\mathbb{E}_{x_S \sim X_S} \mbox{log}(D_{T}(G_{ST}(x_S))) + \\
	\mathbb{E}_{x_T \sim X_T} \mbox{log}(1 - D_{T}(x_T))
\end{multline}

In addition, Cycle-GAN \cite{CycleGAN} introduced a cycle consistency loss $\mathcal{L}_{CYC}$. Cycle consistency enforces the style-transfer network to remain the image content without hallucinating new objects. Using a L1 reconstruction loss, cycle consistency is calculated as shown in Equation \ref{eq:GAN3} for both transformation directions:

\begin{multline}
	\label{eq:GAN3}
	\mathcal{L}_{CYC}(G_{ST}, G_{TS}, x_S, x_T) = \\
	\mathbb{E}_{x_S \sim X_S} \| G_{TS}(G_{ST}(x_S)) - x_S\|_1 + \\
	\mathbb{E}_{x_T \sim X_T} \| G_{ST}(G_{TS}(x_T)) - x_T\|_1
\end{multline}

Introduced by CyCADA \cite{CyCADA}, semantic consistency is used to retain semantic information through the style-transformation. Cycle-GAN often aligns data distributions by e.g. generating trees in sky areas, which leads to decreased performance of downstream tasks on stylized images. The authors of \cite{menke2022} observe that semantic consistency is generally a well-suited regularization mechanism for style-transfer tasks, independent of the downstream task. Using a pre-trained semantic segmentation network $F$ trained on the source domain Equation \ref{eq:GAN4} shows the Kullback-Leibler divergence between predictions of the semantic consistency network on source and stylized source samples. 

\begin{multline}
	\label{eq:GAN4}
	\mathcal{L}_{sem}(G_{ST}, F, x_S) = \\
	\mathbb{E}_{x_S \sim X_S} KL(F(G_{ST}(x_S))||F(x_S))
\end{multline}

For training the full style-transfer model, all loss functions, including GAN losses of both generator and discriminator pairs, cycle-consistency loss, and semantic consistency loss, are weighted by factors $a_1$ to $a_4$ we derive from CyCADA \cite{CyCADA}. We sum all weighted losses for the final loss as shown in Equation \ref{eq:GAN5}:

\begin{equation}
	\label{eq:GAN5}
	\mathcal{L}_{total} = a_1\mathcal{L}_{GAN}^{ST} + a_2\mathcal{L}_{GAN}^{TS} + a_3\mathcal{L}_{CYC} + a_4\mathcal{L}_{sem}
\end{equation}

\subsection{Attention Map Construction}
\label{sec:sampling}
In this Section we describe how we construct our attention maps for training AWADA. Object detector models typically create object proposals separating foreground and background regions. We sample proposals for source domain images from a pre-trained object detector trained on the source domain. We sample proposals from an object detector trained on CyCADA* stylized source images for the target domain to create more accurate attention maps for the target domain than by training on the source domain directly. Compared to perfect predictions, proposals additionally introduce some context and hard negatives into training, which we found to be in fact beneficial in preliminary experiments. From object proposals, we create attention maps, highlighting foreground regions and masking out background regions. 

\begin{equation}
	\label{eq:sampling}
	a_f(u,v) = \mbox{hard}(S^c(u,v)) = 
	\begin{cases}
		1 &\text{if $\left|S^c(u,v)\right|$ \textgreater \hphantom{} 0}\\
		0 &\text{otherwise} 
	\end{cases}
\end{equation}

Attention maps are created on input image resolution using the hard accumulation method $f$. Our method receives a set $S^c(u,v)$ of proposal confidences $\geq c$ for proposals at position (u,v) and outputs the attention map value based on Equation \ref{eq:sampling}. For the source domain, we construct $N_S$ attention maps $A_{f,S} = \{a_{f,S}^i\}_{i = 1}^{N_S}$. For the target domain, we construct $N_T$ attention maps $A_{f,T} = \{a_{f,T}^i\}_{i = 1}^{N_T}$ the same size as the original patch.

\subsection{Attention-Weighting Module}

As seen in Figure \ref{fig:Framework}, we use our proposed Attention-Weighting Modules (AWM) to re-weight the style-transfer network loss functions in AWADA. \\
Typically, loss functions are averaged over all spatial loss pixels. In our AWMs we move this averaging behind the weighting of the per-pixel loss $\mathcal{L}_{pix}$ with the corresponding attention map $a_{f,d}$ of domain $d$.

\begin{equation}
	\mathcal{L}_{w}(\mathcal{L}_{pix}, a_{f,d}) = \dfrac{1}{WH}\sum_{u=1}^{W} \sum_{v=1}^{H} \mathcal{L}_{pix}(u,v) \cdot a_{f,d}(u,v)
\end{equation}

From the attention map associated with the input image, we crop the same region we used to crop a random input image patch. We then resize the cropped attention map to be the same size as the pixel-wise loss map.\\
We multiply the per-pixel loss $\mathcal{L}_{pix}$ with the associated cropped and resized attention map $a_{f,d}$. Thus, only regions highlighted in $a_{f,d}$ remain in the loss and background is multiplied with zero. Afterwards we calculate the mean of the weighted loss map of size $W$x$H$ to form the final weighted mean loss $\mathcal{L}_{w}$. This weighted loss is used in training to perform the model weight update.\\
Our generators should produce high quality foreground object regions, as we mask out background regions which are less relevant for the final detection task. We do not weight the cycle and semantic constsiency losses, as they are responsible for regularizing the whole GAN training. Further analyses are performed on this in the ablation studies.

\begin{table*}[t]
	\centering
	\caption{Results of detectors trained on the Cityscapes Dataset and evaluated on Foggy-Cityscapes for adverse weather and on BDD100k for cross-camera adaptation. We report mean and standard deviation of nine experiments each.}
	\label{table:City2Foggy}
	\begin{tabular}{c|c|cccccccc|c}
		\specialrule{1.2pt}{1pt}{1pt}
		\textbf{Method} & \textbf{Dataset} &\textbf{Person} & \textbf{Rider}& \textbf{Car}& \textbf{Truck}& \textbf{Bus}& \textbf{Train}& \textbf{Mcycle}& \textbf{Bicycle}& \textbf{mAP}\\
		\specialrule{1.2pt}{1pt}{1pt}
		Baseline&Foggy&32.9&38.7&40.7&15.7&25.4&2.5&20.1&40.3&27.0\\
		\hline
		iFAN \cite{iFAN}&Foggy&32.6&40.0&48.5&27.9&45.5&31.7&22.8&33.0&35.3\\
		UDA \cite{UDA}& Foggy&38.2&42.1&55.6&25.9&43.5&27.6&33.5&39.2&38.2\\
		FFDA \cite{FFDA}&Foggy&33.8&48.3&50.7&26.6&49.2&39.4&35.8&36.8&40.1\\
		UMT \cite{UMT} & Foggy&33.0&46.7&48.6&34.1&\textbf{56.5}&46.8&30.4&37.3&41.7\\
		CGD \cite{li2021category}&Foggy&38.0&47.4&53.1&34.2&47.5&41.1&38.3&38.9&42.3\\
		ILLUME \cite{ILLUME}& Foggy &35.8&45.1&54.3&\textbf{34.5}&49.7&\textbf{50.3}&\textbf{38.7}&42.0&43.8\\
		CyCADA* \cite{menke2022}&Foggy&42.1&49.1&53.8&21.8&41.7&26.0&29.4&45.7&38.7 $\pm$ 0.96  \\
		AWADA (Ours)&Foggy&\textbf{44.7}&\textbf{52.7}&\textbf{60.9}&28.7&51.0&36.2&36.0&\textbf{48.5}&\textbf{44.8 $\pm$ 0.65}\\
		\hline
		Oracle&Foggy&50.2&54.8&66.4&33.3&57.0&41.0&36.2&51.4&48.8\\
		\specialrule{1.2pt}{1pt}{1pt}
		Baseline&BDD100k&34.2&24.2&53.7&15.0&14.3&-&12.3&19.5&24.8\\
		\hline
		AFL \cite{AFL} & BDD100k &32.4&32.6&50.4&20.6&23.4&-&18.9&25.0&29.0\\
		PDA \cite{ProgressiveDomainAdaptation}& BDD100k &37.6&32.9&51.8&19.3&23.7&-&16.1&25.3&29.5\\
		ILLUME \cite{ILLUME} & BDD100k &33.2&20.5&47.8&\textbf{20.8}&\textbf{33.8}&-&\textbf{24.4}&26.7&29.6\\
		CyCADA* \cite{menke2022}&BDD100k&40.4&33.1&55.8&15.0&18.1&-&17.4&29.2&29.8 $\pm$ 0.90\\
		AWADA (Ours)&BDD100k&\textbf{41.5}&\textbf{34.2}&\textbf{56.0}&18.7&20.0&-&20.4&\textbf{29.7}&\textbf{31.5 $\pm$ 0.36}\\
		\hline
		Oracle&BDD100k&52.7&42.8&73.0&53.8&52.5&-&36.4&39.5&50.1\\
		\specialrule{1.2pt}{1pt}{1pt}
	\end{tabular}
\end{table*}

\section{Experiments}
In this section, we evaluate AWADA on a variety of unsupervised domain adaptation benchmarks and compare to the state of the art. Finally, we conduct ablation studies to verify the benefits of each component of AWADA separately.

\subsection{Experimental Setup}
We evaluate AWADA on three automotive benchmarks for object detection: Synthetic-to-Real, Adverse Weather and Cross-Camera domain..
\subsubsection{Synthetic-to-Real.} Recorded from the video game GTA5, we choose sim10k \cite{sim10k} as our source dataset and Cityscapes \cite{Cityscapes} as our target dataset for synthetic-to-real adaptation. As Cityscapes is made for semantic segmentation, we use the tightest rectangle of its instance masks to create bounding box labels following \cite{DAFasterRCNN}. The sim10k dataset contains 10,000 annotated images. We picked only 6,591 day images, because Cityscapes does not contain any night images following \cite{menke2022}. We find that night images decrease the style-transfer performance. In this setting we only consider the car class. We do not use labels of Cityscapes except for evaluation on the 500 validation images. This benchmark is particularly interesting because sim10k and Cityscapes have approximately the same image resolution and different object instance appearances and is therefore a challenging task for domain adaptation.
\subsubsection{Cross-Camera.} 
In the setting of cross-camera domain adaptation we choose Cityscapes as our source and BDD100k \cite{bdd100k} as our target domain. BDD100k daytime set contains 36,728 images with bounding box annotations, which we choose because Cityscapes has no night images. We evaluate the cross-camera adaptation on seven common classes of both datasets. This benchmark imposes challenges by its high diversity and inter-frame domain gap.

\subsubsection{Adverse Weather.} For adverse weather adaptation we choose Cityscapes \cite{Cityscapes} as the source domain and Foggy-Cityscapes \cite{FoggyCityscapes} as the target domain. Foggy Cityscapes is created by a fog simulation run on original Cityscapes images. We choose only the highest fog level with a sight of view of about 150 meters, matching the setting in \cite{SeekingSimilarities}. Cityscapes contains 2975 labeled training images with eight different road user classes. Foggy-Cityscapes validation set contains 500 images.

\subsubsection{Implementation Details.}
We choose Faster-RCNN as our object detector as this is the default in most other work \cite{DAFasterRCNN,StrongWeak,ProgressiveDomainAdaptation,SeekingSimilarities,iFAN,CDN}. We resize all images to 600px shortest side and batch-size one. With a VGG16 \cite{VGG} backbone, we train for 30 epochs using Adam optimizer and a learning rate of 1e-5. We report results using mean average precision (mAP) with an intersection over union (IoU) threshold of 0.5.\\
We train AWADA for 200 epochs with a batch size of two. For all other hyperparameters, we follow the protocol of CyCADA* \cite{menke2022}. In attention map creation, we take proposals with a minimum confidence score of 0.5 into account. To save computation time, we pre-compute the attention maps for each image of the source and target domain. As GAN training is instable, we report results as an average of nine experiments, training three GANs and evaluate each using three detectors for each of our reported results using different random seeds.\\
We present results in comparison with a baseline trained directly on the source domain while evaluating on the target domain. As an upper bound, we additionally report results from training on the labeled target domain as an oracle. 

\subsection{Experimental Results}
Our experiments first compare AWADA to the current state-of-the-art methods in unsupervised domain adaptation. Table \ref{table:City2Foggy} shows results of detectors trained on an AWADA stylized Cityscapes dataset and evaluated on the Foggy-Cityscapes dataset for adverse weather and on BDD100k dataset for cross-camera adaptation. AWADA outperforms all current SOTA methods on both benchmarks. Both benchmarks are highly complex and show that AWADA is capable of improving detection performance on different automotive relevant classes. As we do not modify the object detector network, AWADA can be combined with approaches such as ILLUME or CGD to likely further improve object detection performance. \\
In Table \ref{table:GTA2City2}, we report results of AWADA on synthetic-to-real adaptation and show, that AWADA is on paar with other methods on the synthetic-to-real benchmark. From the repetition of our experiments we see the increased stability of AWADA compared to CyCADA*. Unfortunately we cannot conduct such a comparison to other methods, since the corresponding results have not been reported. \\
AWADA requires a total of 88 hours trained on a single V100 GPU. As we pre-compute the attention maps, the remaining AWADA training requires 40 GPU hours, which is only about 10 hours longer than a CyCADA* training.

\begin{table}
	\begin{center}
		\caption{Results in the sim10k to Cityscapes synthetic-to-real adaptation benchmark. We report mean and standard deviation of nine experiments each.}
		\label{table:GTA2City2} 
		\begin{tabular}{c|c}
			\specialrule{1.2pt}{1pt}{1pt}
			\textbf{Method}& \textbf{car AP}\\
			\specialrule{1.2pt}{1pt}{1pt}
			Baseline& 34.7\\
			\hline
			iFAN \cite{iFAN} & 46.9\\
			UDA \cite{UDA} & 52.3 \\
			SCAN \cite{li2022scan}& 52.6 \\
			ILLUME \cite{ILLUME} & 53.1 \\
			CyCADA* \cite{menke2022}& 51.9 $\pm$ 0.69\\
			AWADA (Ours)& \textbf{53.2 $\pm$ 0.29} \\
			\hline
			Oracle& 71.1\\
			\specialrule{1.2pt}{1pt}{1pt}
		\end{tabular}
	\end{center}
\end{table}

\subsection{Ablation Studies}
In this section, we conduct ablation studies on our proposed AWADA framework. We first show experiments on different types of loss functions we apply our Attention-Weighting Modules (AWM) to. Afterwards we show the influence of our attention map creation strategy on the object detection performance. At the end we analyze the total runtime of AWADA.

\subsubsection{Attention-Weighting of Losses.}
Our first ablation study targets the choice of loss functions we attach our AWMs to. We conduct experiments in which we re-weight the various loss functions that appear in our AWADA framework. We define four groups representing the discriminator $\mathcal{L}_{disc}$, the generator $\mathcal{L}_{gen}$, the cycle-consistency $\mathcal{L}_{cyc}$, and the semantic-consistency loss $\mathcal{L}_{sem}$. 

\begin{table}[h]
	\begin{center}
		\caption{Ablation study on the different losses that are weighted with object detection attention maps using our proposed AWM on the sim10k to Cityscapes benchmark. The check mark indicates the application of an AWM to the loss, otherwise the loss remains unweighted.}
		\label{table:abl_weighting}
		\begin{tabular}{cccc|c}
			\specialrule{1.2pt}{1pt}{1pt}
			\textbf{$\mathcal{L}_{disc}$} & \textbf{$\mathcal{L}_{gen}$}  & \textbf{$\mathcal{L}_{cyc}$} & \textbf{$\mathcal{L}_{sem}$} & \textbf{car AP}\\
			\specialrule{1.2pt}{1pt}{1pt}
			-& - & - &- & 51.9 $\pm$ 0.69\\
			\checkmark& - &- &- &50.5 $\pm$ 0.25\\
			-& \checkmark &- & -&52.4 $\pm$ 0.65\\
			\checkmark& \checkmark &- &- &\textbf{53.2 $\pm$ 0.29}\\
			\checkmark& \checkmark & \checkmark & -& 50.5 $\pm$ 1.07\\
			\checkmark& \checkmark & \checkmark & \checkmark & 51.4 $\pm$ 0.72\\
			\specialrule{1.2pt}{1pt}{1pt}
		\end{tabular}
	\end{center}
\end{table}

In Table \ref{table:abl_weighting}, we see that the weighting of only the loss functions connected to the adversarial optimization problem $\mathcal{L}_{disc}$ and $\mathcal{L}_{gen}$ performs the best. In style-transfer, the generators and discriminators compete with each other. Re-weighting only the generator or the discriminator loss leads to a decrease of performance, which is likely to be caused by imbalance of the adversarial optimization problem. This suggests re-weighting these losses equally. Cycle-consistency and semantic consistency are responsible for the supervision and regularization of the generation cycle. In order to balance regularization and focus on foreground object regions we do not re-weight regularization losses, which shows best results. \\
We also tested normalizing the attention-weighted loss, but found results to significantly degrade. This shows that the natural AWADA weighting is most beneficial for training. 

\begin{figure*}[t]
	\centering
	\includegraphics[width=1.0\textwidth]{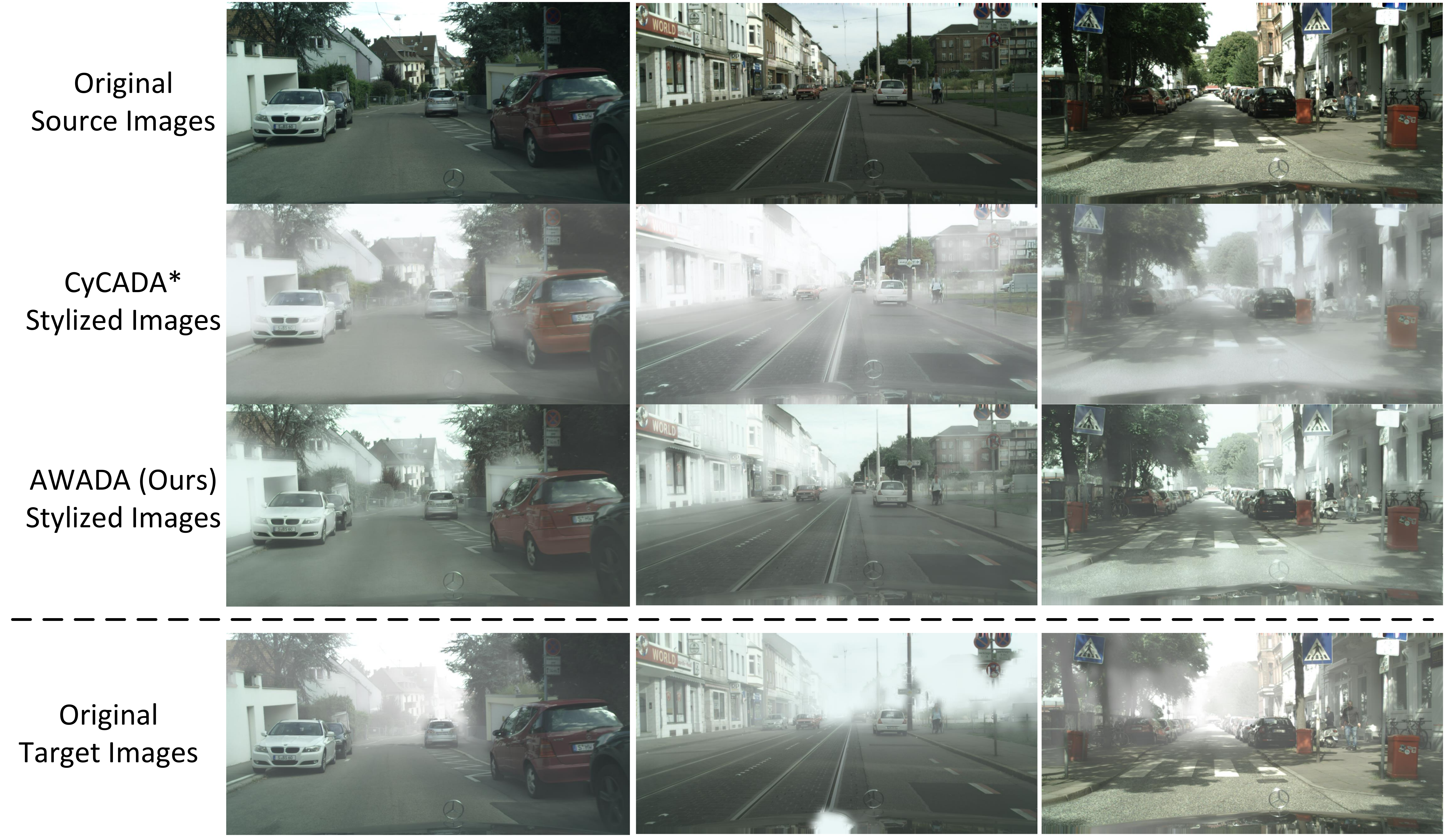}
	\caption{Qualitative Results of the adverse weather Cityscapes \cite{Cityscapes} to Foggy-Cityscapes \cite{FoggyCityscapes} Benchmark. It can be seen, that CyCADA* produces very strong and deep fog even for close image regions. In contrast, our AWADA framework produces more realistic fog because of the regularization based on foreground focused transformation we apply.}
	\label{fig:foggy_qualitative}
\end{figure*}

\subsubsection{Attention Map Construction Method.}
\label{sec:other}
Next we analyze the quality of our constructed attention maps and the influence on the detection performance. Our default AWADA implementation uses object detector proposals to construct attention maps. \\
At first, we evaluate the influence of our fuzzy, imperfect attention maps constructed from proposals. Perfect attention maps could be obtained when we use ground truth semantic segmentation masks or bounding boxes. As shown in Table \ref{table:other}, using perfect attention maps from bounding boxes or segmentation degrades results. We also compared inflating GT boxes by 20\%, which improved results slightly. We hypothesize that adding context around objects is beneficial for the downstream detection training.

\begin{table}[h]
	\begin{center}
		\caption{Ablation study on different ground truth attention map creation methods on the sim10k to Cityscapes benchmark.}
		\label{table:other}
		\begin{tabular}{c|ccc}
			\specialrule{1.2pt}{1pt}{1pt}
			\textbf{Method}  & \textbf{GTA2City} & \textbf{City2Foggy} & \textbf{City2BDD}\\
			\specialrule{1.2pt}{1pt}{1pt}
			CyCADA* & 51.9 $\pm$ 0.69 & 38.7 $\pm$ 0.96& 29.8 $\pm$ 0.90\\
			\specialrule{1.2pt}{1pt}{1pt}
			GT Masks & 50.4 $\pm$ 0.46 & 38.8 $\pm$ 1.07& - \\
			GT Boxes & 51.9 $\pm$ 0.82 & 40.9 $\pm$ 1.36&29.9 $\pm$ 0.55\\
			GT Inflate & 52.6 $\pm$ 0.59 &42.0 $\pm$ 0.98 &29.6 $\pm$ 0.52\\
			\specialrule{1.2pt}{1pt}{1pt}
			AWADA & \textbf{53.2 $\pm$ 0.29} & 44.8 $\pm$ 0.65& \textbf{31.5 $\pm$ 0.36}\\
			\specialrule{1.2pt}{1pt}{1pt}
		\end{tabular}
	\end{center}
\end{table}

Table \ref{table:other2} additionally shows results of using 10-50\% of randomly marking pixels as foreground. This is an ablation and simplistic application of our loss-masking framework, where we mask out a fixed percentage of random pixels instead of object boxes. This method works surprisingly well in some scenarios, yet results are sensitive to the choice of the randomness hyperparameter and increased instability. AWADA results however have been stable across all datasets using the same set of hyperparameters, which indicates further optimization potential of AWADA compared to random results. We also tried combining AWADA with random, with results in between both individual methods, but again increased instability.
Therefore despite its simplicity, we cannot generally recommend using the random method and prefer AWADA for more stable results. \\
If we simply reduce the adversarial loss weights in AWADA without using AWMs, the training collapses, therefore the improvement in AWADA does not come from reduced loss alone. \\
We also tried different proposal accumulation functions $f$ as mean, median or max which did not improve the results. Therefore we stick to our hard accumulation function. Further results are supplied in the supplementary material.

\begin{table}[h]
	\begin{center}
		\caption{Ablation study on different random attention map creation methods on the sim10k to Cityscapes benchmark.}
		\label{table:other2}
		\begin{tabular}{c|ccc}
			\specialrule{1.2pt}{1pt}{1pt}
			\textbf{Method}  & \textbf{GTA2City} & \textbf{City2Foggy} & \textbf{City2BDD}\\
			\specialrule{1.2pt}{1pt}{1pt}
			CyCADA* & 51.9 $\pm$ 0.69 & 38.7 $\pm$ 0.96& 29.8 $\pm$ 0.90\\
			\specialrule{1.2pt}{1pt}{1pt}
			Rnd. 10 \% & 52.9 $\pm$ 0.70 & 42.7 $\pm$ 0.62&26.8 $\pm$ 0.71\\
			Rnd. 30 \% & 52.8 $\pm$ 0.68 & 45.7 $\pm$ 0.42&29.7 $\pm$ 0.24\\
			Rnd. 50 \% & 51.5 $\pm$ 0.47 &\textbf{46.2 $\pm$ 0.61}&30.3 $\pm$ 0.48\\
			\specialrule{1.2pt}{1pt}{1pt}
			AWADA & \textbf{53.2 $\pm$ 0.29} & 44.8 $\pm$ 0.65& \textbf{31.5 $\pm$ 0.36}\\
			\specialrule{1.2pt}{1pt}{1pt}
		\end{tabular}
	\end{center}
\end{table}

\section{Conclusion}

We propose AWADA, an Attention-Weighted Adversarial Domain Adaptation framework. We direct our style-transfer training to focus the adversarial optimization problem of generators and discriminators on foreground object regions. By creating attention maps from foreground object proposals, we for the first create a feedback loop between detection and transformation tasks in GAN-based style transfer. AWADA does not only improve object detection performance, it also stabilizes the training of the GAN network. \\
Experiments and ablation studies show that our AWADA framework outperforms current unsupervised domain adaptation methods on synthetic-to-real, adverse weather and cross-camera automotive object detection benchmarks and provide a more stable and regularized image-to-image style-transfer method.

{\small
\bibliographystyle{ieee_fullname}
\bibliography{egbib}
}

\end{document}